\documentclass[10pt,twocolumn,letterpaper]{article}

\usepackage[pagenumbers]{cvpr} % To force page numbers, e.g. for an arXiv version

\makeatletter
\@namedef{ver@everyshi.sty}{}
\makeatother
\usepackage{tikz}

\usepackage[accsupp]{axessibility}
\usepackage{amsmath}
\usepackage{amssymb}
\usepackage{booktabs}
\usepackage{bbm}
\usepackage{adjustbox}
\usepackage{graphicx}
\usepackage{tabularx,ragged2e,booktabs}
\usepackage{algorithm, algpseudocode}
\newcommand{\xx}{\mathbf{x}}
\newcommand{\cc}{\mathbf{c}}
\newcommand{\yy}{\mathbf{y}}

\newcommand{\CC}{\mathbf{C}}
\newcommand{\tyy}{\tilde{\mathbf{y}}}
\newcommand{\hyy}{\hat{\mathbf{y}}}
\newcommand{\ty}{\tilde{y}}

\newcommand{\tCC}{\tilde{\mathbf{C}}}

\newcommand{\mL}{\mathcal{L}}

\newcommand{\issue}[1]{\vspace{0.1em}\noindent {\textbf{#1 \hspace{0.2em}}}}
\usepackage[pagebackref,breaklinks,colorlinks]{hyperref}
\usepackage[capitalize]{cleveref}
\crefname{section}{Sec.}{Secs.}
\Crefname{section}{Section}{Sections}
\Crefname{table}{Table}{Tables}
\crefname{table}{Tab.}{Tabs.}

\pdfoutput=1

\def\cvprPaperID{4790} % *** Enter the CVPR Paper ID here
\def\confName{CVPR}
\def\confYear{2023}

\begin{document}

\title{Semi-Supervised Domain Adaptation with Source Label Adaptation}

\author{Yu-Chu Yu\qquad Hsuan-Tien Lin\\
National Taiwan University\\
% Taipei, Taiwan\\ 
{\{\tt\small r09922104, htlin\}@csie.ntu.edu.tw}
% \and
}
\maketitle

%%%%%%%%% ABSTRACT
\begin{abstract}
   Semi-Supervised Domain Adaptation (SSDA) involves learning to classify unseen target data with a few labeled and lots of unlabeled target data, along with many labeled source data from a related domain. Current SSDA approaches usually aim at aligning the target data to the labeled source data with feature space mapping and pseudo-label assignments. Nevertheless, such a source-oriented model can sometimes align the target data to source data of the wrong classes, degrading the classification performance. This paper presents a novel source-adaptive paradigm that adapts the source data to match the target data. Our key idea is to view the source data as a noisily-labeled version of the ideal target data. Then, we propose an SSDA model that cleans up the label noise dynamically with the help of a robust cleaner component designed from the target perspective. Since the paradigm is very different from the core ideas behind existing SSDA approaches, our proposed model can be easily coupled with them to improve their performance. Empirical results on two state-of-the-art SSDA approaches demonstrate that the proposed model effectively cleans up the noise within the source labels and exhibits superior performance over those approaches across benchmark datasets. Our code is available at \url{https://github.com/chu0802/SLA}.

\end{abstract}

% %%%%%%%%% BODY TEXT
% \section{Introduction}
% \label{sec:intro}
% \input{Section/Introduction}
% %------------------------------------------------------------------------
% \section{Formatting your paper}
% \label{sec:formatting}
% \input{Section/Second}
% %------------------------------------------------------------------------
% \section{Final copy}
% \input{Section/Third}
\section{Introduction}
\label{sec:intro}
% Domain Adaptation is a generalized machine learning problem with a specific assumption. In the scenario, we assume that the training data and test data are drawn from similar, but different domains, named source domain and target domain, respectively. 

\begin{figure}[t]
  \centering
  \includegraphics[width=0.91\linewidth]{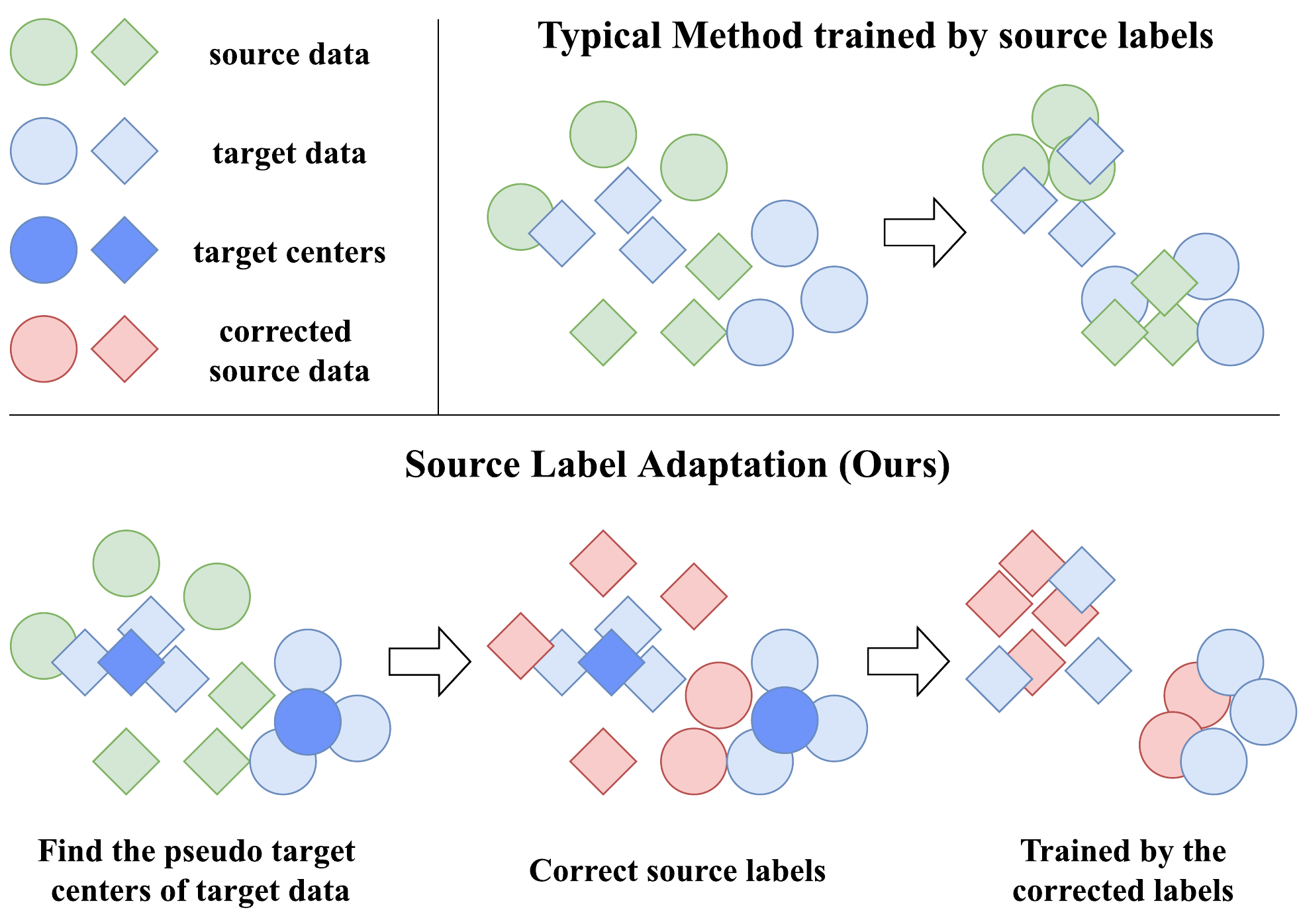}
   \caption{\textbf{Top.} Training the model with the original source labels might lead to the misalignment of the target data. \textbf{Bottom.} After cleaning up the noisy source labels with our SLA framework, the  target data can be aligned with the correct classes.}
   
   \label{fig:intro}
\end{figure}

% Traditional machine learning problems assume that training and test data are drawn from the same distribution. Based on this assumption, numerous techniques have been proven to be theoretically and practically effective in various fields \cite{ben2010}. 
% \htcomment{
% It is a bit strange to use this citation to support (the success of) "traditional machine learning." Also, this opening sentence is long but not very exciting.....
% }
% \yccomment{
% I think we can remove the first sentence, directly discussing domain adaptation.
% }
Domain Adaptation (DA) focuses on a general machine learning scenario where training and test data may originate from two related but distinct domains: the source domain and the target domain. 
% \htcomment{
% Domain Adaptation (DA) focuses on a general machine learning scenario where training and test data may originate from two related but distinct domains: the source domain and the target domain.
% }
Many works have extensively studied unsupervised DA (UDA), where labels in the target domain cannot be accessed, from both theoretical~\cite{ben2010, daot2016, invariant2019} and algorithmic \cite{dann2015, long2016unsupervised, kang2019contrastive, saito2018maximum, Na_2021_CVPR, zhao2021reducing} perspectives.
% \htcomment{
% Many works have extensively studied unsupervised DA, where no labels in the target domain can be accessed, from theoretical~\cite{ben2010, invariant2019} and algorithmic \cite{dann2015, long2016unsupervised, kang2019contrastive, saito2018maximum, Na_2021_CVPR, zhao2021reducing} angles.
% }
% \htcomment{
% Many works have extensively studied unsupervised DA where labels in the target domain are not accessible.
% }
% There are also several rigorous theories that have been proposed to guarantee the upper and lower bounds of a domain adaptation algorithm's performance \cite{}. 
% \htcomment{
% It is not clear that we need to talk about theory here, as this work is far from theoretical, and mentioning this (so early) gives people the expectation that you will talk about theory too.
% Or you can combine this sentence with the previous one, say something like
% from theoretical~\cite{...} and algorithmic~\cite{} angles.
% }
% Nevertheless, achieving considerable progress with the help of only a few target labels is far simpler than designing an intricate algorithm in Domain Adaptation. 
Recently, Semi-Supervised Domain Adaptation (SSDA), another DA setting that allows access to a few target labels, has received more research attention because it is a simple yet realistic setting for application needs.
% \htcomment{
% Recently, Semi-Supervised Domain Adaptation (SSDA), another DA setting that allows accessing a few target labels, is gaining more research attention for being simple yet realistic for application needs.
% }
% \htcomment{
% Given that you did not talk about UDA before this sentence, the readers do not have a sense of "simpler than" here. I think you can be direct and just say that SSDA is a simple yet realistic setting in DA that allows a few target labels.
% }
% Since getting a small number of target labels should not be a tough challenge, Semi-Supervised Domain Adaptation (SSDA) has attract more attention in recent years.

The most na\"{i}ve strategy for SSDA, commonly known as S+T \cite{fsl2018, mme2019}, aims to train a model using the source data and labeled target data with a standard cross entropy loss. This strategy often suffers from a well-known \textit{domain shift} issue, which stems from the gap between different data distributions. 
% \htcomment{
% The most naive strategy for SSDA, commonly known as S+T \cite{fsl2018, mme2019}, aims at training a model using the source data and labeled target data with a standard cross entropy loss. 
% }
% Based on the strategy, several state-of-the-art algorithms attempt to further explore the better usage of unlabeled target data so that the target data distribution can be appropriately aligned with the source data distribution. 
% \htcomment{
% What's the difference between "method" and "strategy"? Usually good to be consistent.
% }
% \htcomment{
% The naive strategy is extended to several state-of-the-art algorithms that explore better use of the unlabeled target data. 
% }
To address this issue, many state-of-the-art algorithms attempt to explore better use of the unlabeled target data so that the target distribution can be aligned with the source distribution. Recently, several Semi-Supervised Learning (SSL) algorithms have been applied for SSDA \cite{mme2019, cdac2021, mcl2022} to regularize the unlabeled data, such as entropy minimization \cite{ent2004}, pseudo-labeling \cite{pseudo2013, fixmatch2020} and consistency regularization \cite{consistency2014, fixmatch2020}. These classic source-oriented strategies have prevailed for a long time. 
However, these algorithms typically require the target data to closely match some semantically similar source data in the feature space. Therefore, if the S+T space has been misaligned, it can be challenging to recover from the misalignment, as illustrated in Figure~\ref{fig:intro}.

\begin{figure}[t]
  \centering
  \includegraphics[width=0.91\linewidth]{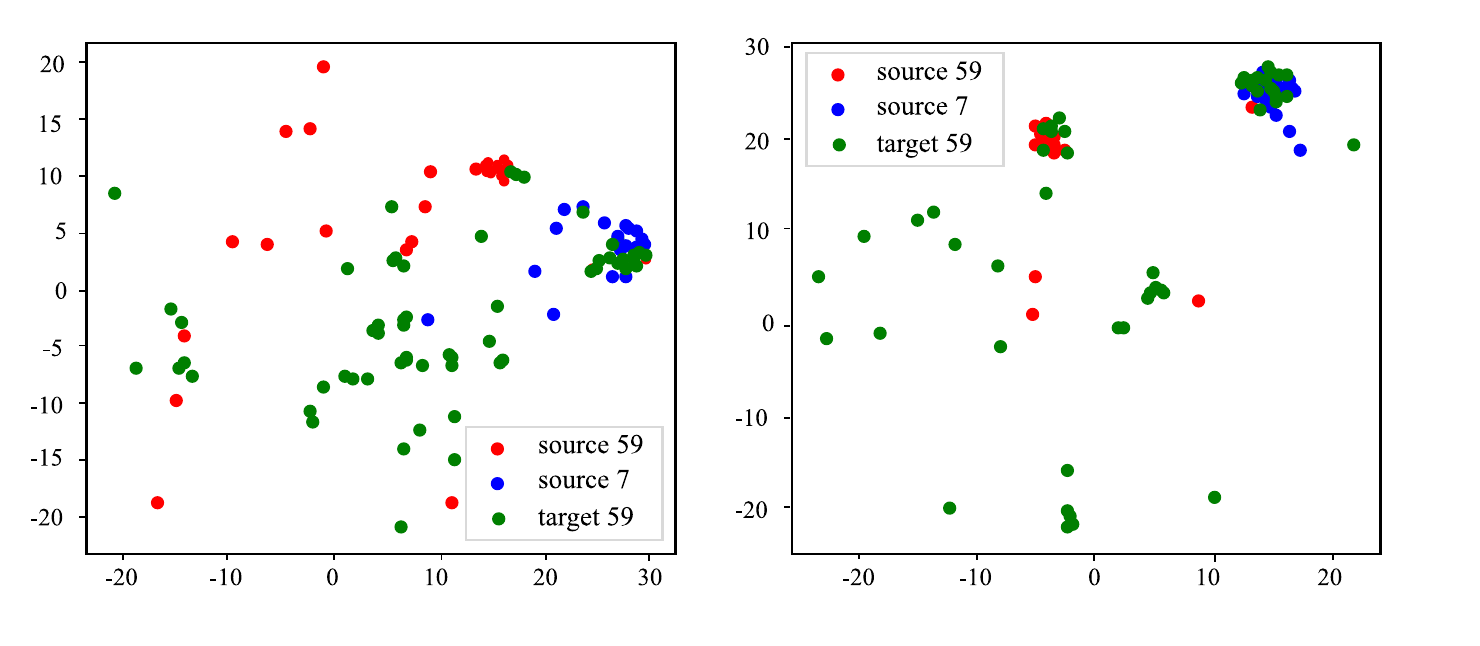}
   \caption{Feature visualizations with t-SNE for an example of the misalignment on the \textit{Office-Home} A $\to$ C dataset with ResNet34. The model is trained by S+T. \textbf{Left}: 0-th iteration. \textbf{Right}: 5000-th iteration . We observe that the misalignment has already happened at a very early stage. Guided by source labels and a few target labels, a portion of the target data from the 59th class misaligns with the source data from the 7th class.}
   
   \label{fig:misalign}
\end{figure}

We take a deeper look into a specific example from the \textit{Office-Home} dataset \cite{venkateswara2017deep} to confirm the abovementioned issue. Figure~\ref{fig:misalign} visualizes the feature space trained by S+T using t-SNE \cite{chan2019gpu}. We observed that the misalignment between the source and the target data has happened at a very early stage. For instance, in the beginning, a portion of the target data from the 59th class is close to the source data from the 7th class. Since we only have access to source labels and a few target labels, without proper guidance from enough target labels, such misalignment becomes more severe after being trained by S+T. Table~\ref{tab:misalign} shows the partial confusion matrix of S+T. Roughly 40\% of the target data in the 59th class is mispredicted to the 7th class, and only around 20\% of the data is classified correctly.
% Figure~\ref{fig:misalign} visualizes the \textit{ImageNet} \cite{imagenet} pre-trained space and the feature space trained by S+T using t-SNE \cite{chan2019gpu}. 
% We observe that over the pre-trained space, a portion of target data in 59th class has already been close to source data in 7th class. Since we only have access to source labels and a few target labels, without proper guidance from enough target labels, such misalignment becomes more severe after being trained by S+T. Table~\ref{tab:misalign} shows the partial confusion matrix of S+T. One-third of the target data in 59th class is mispredicted as 7th class, and only around 20\% of the data is classified correctly.
% \htcomment{
% It is not clear what story you'd want to tell for using ImageNet pre-trained space. People might say "because it is not trained with S nor T, it is natural that the space is not that good. So the "misalignment" is expected.

% Telling a good story from S+T since sufficient for me.
% }
% \htcomment{
% Can we say that S+T "overfits" (maybe a bad name, but maybe a similar term) to S?
% }
        
From the case study above, we argue that relying on the source labels like S+T can misguide the model to learn wrong classes for some target data. That is, source labels can be viewed as a \textit{noisy} version of the ideal labels for target classification.
% \htcomment{
% From the case study above, we argue that relying on the source labels like S+T can misguide the model to learning wrong classes for some target data. That is, source labels can be viewed as a \textit{noisy} version of the ideal labels for target classification. 
% }
% \htcomment{
% Classification under noisy labels, or noisy label learning ..., is not a novel problem in machine learning...do some literature reviews
% }
Based on the argument, the setting of SSDA is more like a Noisy Label Learning (NLL) problem, with a massive amount of noisy labels (source labels) and a small number of clean labels (target labels). 

Learning with noisy labels is a widely studied machine learning problem. A popular solution is to clean up the noisy labels with the help of another model, which can also be referred to as label correction \cite{wang2022proselflc}. 
% By cleaning up the label noise within the source labels, the feature space should be closer to the ideal target feature space.
To approach Domain Adaptation as an NLL problem, we borrow the idea from label correction and propose a Source Label Adaptation (SLA) framework, as shown in Figure~\ref{fig:intro}. We construct a label adaptation component that provides the view from the target data and dynamically cleans up the noisy source labels at each iteration. Unlike other earlier works that study how to leverage the unlabeled target data, we mainly investigate how to train the source data with the adapted labels to better suit the ideal target space. This source-adaptive paradigm is entirely orthogonal to the core ideas behind existing SSDA algorithms. Thus, we can combine our framework with those algorithms to get superior results. We summarize our contributions as follows.

\begin{itemize}
    \item We argue that the classic source-oriented methods might still suffer from the biased feature space derived from S+T. To escape this predicament, we propose adapting the source data to the target space by modifying the original source labels.
    % \htcomment{Say more positively about *your* insight, not just saying that existing paradigms are problematic}
    \item We address DA as a particular case of NLL problems and present a novel source-adaptive paradigm. Our SLA framework can be easily coupled with other existing algorithms to boost their performance. 
    %Thus, the adaptation can be bi-directional, further enhancing performance.
    % that cleans up the source labels to better fit the target space.
    % \htcomment{You want to say "noisy learning, but not the typical noisy learning in the literature"}
    % \yccomment{Is it okay to state that it's a special case of NLL?}
    % \htcomment{highlight novelty a bit more}
    \item We demonstrate the usefulness of our proposed SLA framework when coupled with state-of-the-art SSDA algorithms. The framework significantly improved existing algorithms on two major benchmarks, inspiring a new direction for solving DA problems.
    % \htcomment{
    %   We demonstrate the usefulness of our proposed SLA framework when coupled with state-of-the-art SSDA algorithms. The framework significantly improved existing algorithms on several benchmarks and achieved new state-of-the-art results.
    % }
\end{itemize}

% To approach Domain Adaptation as a Noisy Label Learning problem, we propose a Label Correction (LC) framework for Domain Adaptation, as Figure~\ref{fig:intro} shows. We construct a label correction model that provides the view from target data, and dynamically corrects the source labels at each iteration. Unlike other earlier efforts that study how to regularize the unlabeled data, we mainly investigate how to train source data with the corrected labels to suit the ideal target space better. Thus, our framework can be combined with other strategies to get superior results. We summarize our contributions as follows.
% \begin{itemize}
%     \item We provide a new perspective on resolving domain shift by addressing Domain Adaptation as a Noisy Label Learning problem.
%     \item We propose a new framework, Label Correction (LC) for Domain Adaptation, to solve the issue.
%     \item We apply our framework to other state-of-the-art SSDA algorithms. The empirical results on several benchmarks indicate that our framework can significantly improve these works.
% \end{itemize}

\begin{table}
\centering
\scalebox{0.9}{
    \begin{tabular}{lcccc}
    
    \toprule
    True\textbackslash{}Pred & Class 7 & Class 59 & Class 41 & Others \\ \midrule
    Class 59                 & 38.5\%  & 19.8\%   & 13.5\%   & 28.2\%\\
    \bottomrule
    \end{tabular}
    }
    \caption{A partial confusion matrix of S+T on the 3-shot \textit{Office-Home} A $\to$ C dataset with ResNet34. 
    % About 40\% of the target data from the 59th class is wrongly classified into the 7th class. Only about 20\% of the data is predicted correctly.
    }
    \label{tab:misalign}
\end{table}

\begin{figure*}[ht]
    \centering
    \includegraphics[width=0.91\linewidth]{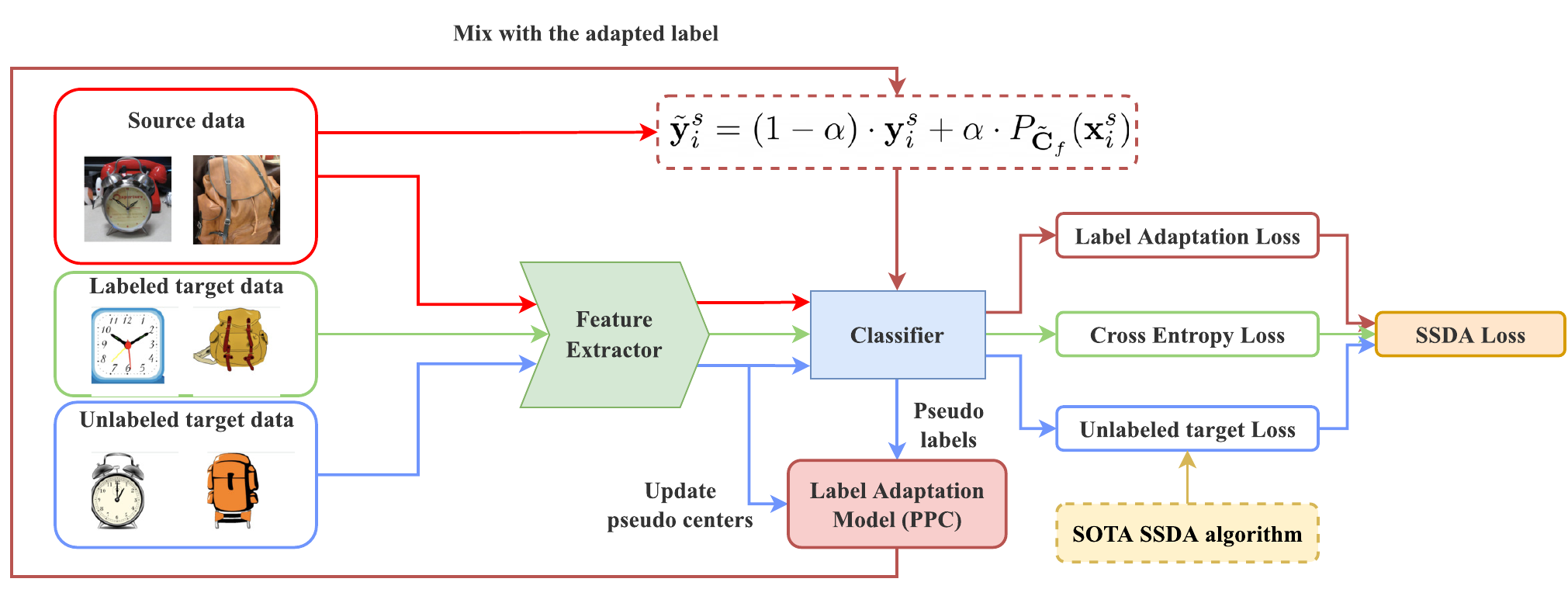}
    \caption{An overview of our proposed framework, Source Label Adaptation for SSDA. For source data, we adapt the original source labels to better fit the target feature space by the Protonet with Pseudo Centers (PPC) and calculate the label adaptation loss. For labeled target data, we train it with a standard cross entropy loss. We can apply a state-of-the-art algorithm to derive the unlabeled target loss for unlabeled data. For every specific interval $I$, we update the pseudo labels and pseudo centers to get a more reliable label adaptation model.}
    \label{fig:archi}
\end{figure*}

% \section{Problem Setup}
% \label{sec:problem}
% \input{Section/Problem}

\section{Related Work}
\label{sec:related}
\issue{Problem Setup.} DA focuses on a $K$-class classification task with an $m$-dimensional input space $X \subseteq \mathbb{R}^m$ and a set of labels $\{1, 2, \ldots, K\}$. For simplicity, we define a label space $Y$ on the probability simplex $\Delta^K$. A label $y = k \in \{1, 2, \ldots, K\}$ is equivalent to a one-hot encoded vector $\yy \in Y$, where only the $k$-th element is $1$ and the others are $0$. We consider two domains over $X \times Y$, named source domain $D_s$ and target domain $D_t$. In SSDA, we sample an amount of labeled source data $S = \{(\xx_i^s,y_i^s)\}_{i=1}^{|S|}$ from $D_s$, labeled target data $L = \{(\xx_i^\ell, y_i^\ell)\}_{i=1}^{|L|}$ from $D_t$, and unlabeled target data $U = \{\xx_i^u\}_{i=1}^{|U|}$ from the marginal distribution of $D_t$ over $X$. Typically, $|L|$ is considerably smaller than $|S|$ and $|U|$, such as one or three examples per class. Our goal is to train an SSDA model $g$ with $S, L$, and $U$ to perform well on the target domain.

\issue{Semi-Supervised Domain Adaptation (SSDA).} SSDA can be viewed as a relaxed yet realistic version of UDA.
% Obtaining a small number of target labels should not be a tough challenge in real-world applications. Thus, SSDA has attracted more attention in recent years.
% \htcomment{
% SSDA can be viewed as a relaxed yet realistic version of Unsupervised Domain Adaptation (UDA). 
% All other messages are mentioned in Introduction and do not need that much repeating here. You can directly go into reviewing SSDA works.
% }
% SSDA has gained more and more attention since it relaxes the original constraints on Unsupervised Domain Adaptation (UDA) so that we can make great progress with only a modest amount of additional resources. 
% \htcomment{
% Generally you do not want to overclaim this "since". You can say that SSDA can be viewed as a relaxed version of UDA, but do not give people the impression that SSDA works are "taking the easier way out". They are just different settings that address different needs. 
% }
% Due to the similarity to Semi-Supervised Learning (SSL), a special case of domain adaptation problems \cite{sslda2021}, many researchers have recently attempted to apply SSL techniques to overcome SSDA problems.
An SSDA algoirthm usually involves three loss functions:
\begin{equation}
    \mL_{\text{SSDA}} = \mL_s + \mL_\ell + \mL_u
\end{equation}
where $\mL_s$ stands for the loss derived by the source data. $\mL_\ell, \mL_u$ denotes the losses from the labeled and unlabeled target data.
% \begin{equation}
%     \mL_{\text{S+T}} = \frac{1}{|S|} \sum_{i=1}^{|S|} \ell_{\text{ce}}(g(\xx_i^s),\yy_i^s) + \frac{1}{|L|}\sum_{i=1}^{|L|} \ell_{\text{ce}}(g(\xx_i^\ell),\yy_i^\ell)
% \end{equation}
% Here $\ell_{\text{ce}}(\cdot, \cdot)$ measures the cross entropy between two probability distributions. 
As discussed in Section~\ref{sec:intro}, based on S+T, a typical SSDA algorithm usually focuses on designing $\mL_u$ to better align the target data with the source data. Recently, many existing works have borrowed SSL techniques to conquer SSDA because of the problem similarity~\cite{sslda2021}.
% \htcomment{
% Many existing works borrow semi-supervised learning (SSL) techniques to conquer SSDA because of the problem similarity~\cite{sslda2021}.
% }
\cite{mme2019} proposes a variant of entropy minimization \cite{ent2004} to explicitly align the target data with source clusters. 
% \htcomment{
% Don't just list the papers. State their connections, especially to S+T (which matches your discussion later). If you need to organize them by a nice equation (like what you did in the oral), consider switching Problem Formulation and Related Work sections so you have the notations ready.
% }
\cite{decota2020} decomposes SSDA into an SSL and a UDA task. The two different sub-tasks produce pseudo labels respectively, and learn from each other via co-training. \cite{cdac2021} groups target features into clusters by measuring pairwise feature similarity. \cite{mcl2022} utilizes consistency regularization at three different levels to perform domain alignment. Besides, both \cite{cdac2021, mcl2022} apply pseudo labeling with data augmentations \cite{fixmatch2020} to enhance their performance. To the best of our knowledge, all methods listed above mainly explore the usage of unlabeled target data while treating the source data with the most straightforward strategy. In our study, we noticed that source labels could appear noisy from the viewpoint of the target data. We thus developed a source-adaptive framework to gradually adapt the source data to the target space. Since we are addressing a new facet of the issue, our framework can be easily applied to several SSDA algorithms mentioned above, further improving the overall performance.

% and we developed a label adaptation framework to use source data more effectively. Since we are addressing a new facet of the issue, our framework can be easily applied to any state-of-the-art algorithm mentioned above, further improving the overall performance.

\issue{Noisy Label Learning (NLL).} The effectiveness of a machine learning algorithm highly depends on the quality of collected labels.
With regard to the present deep neural network design \cite{resnet2015}, the aforementioned issue could worsen as deep models have the capability to fit the data set in a seemingly random manner, regardless of the quality of the labels \cite{rethinking2016}.%Especially for the current deep neural network architecture \cite{resnet2015}, such problems might get much worse since a deep model can usually arbitrarily fit the dataset even if the labels are random \cite{rethinking2016}. 
To clean the noisy labels, \cite{reed2014} proposes a smoothing mechanism to mix noisy labels with self-prediction. \cite{joint2018} models clean labels as trainable parameters and designs a joint optimization algorithm to alternatively update parameters. \cite{losscorrection2016, nl2014, dualt2020} estimate a transition matrix to correct the corrupted labels. However, learning a global transition matrix usually need a strong assumption of how noisy labels come from, which is difficult to verify in the real-world scenarios \cite{instancedependent2020}. \cite{mlc2019} trains a label correction network in a meta-learning manner to help correct noisy labels. Motivated by \cite{reed2014, mlc2019}, we propose a simple framework that can efficiently build a label adaptation model to correct the noisy source labels. 
%By modifying the source labels, we adapt the noisy source labels to better fit the ideal labels for target classification.

% \textbf{Few-Shot Learning (FSL).} Learning with a limited number of instances to recognize new unseen classes is critical to apply machine learning to more practical applications. Though there is a difference between the goal of FSL and SSDA, as pointed out by \cite{mme2019}, the fundamental idea is to extend models to a new domain or a new class with only few labels for each class. Metric-based methods \cite{match2016, protonet2017, fsl2018, fsl2020, fsl2021} learn an embedding function so that the distance over the metric space can determine the similarity among instances. It turns out that the selection of the representative prototype for each class dramatically influences the performance. In our work, we build a Prototypical Network \cite{protonet2017} as a label adaptation model that can provide the view from target data. Nonetheless, we argue that with only few labels per class, the estimated prototypes in \cite{protonet2017} might be biased towards the ideal case. Fortunately, we can better estimate the ideal prototypes with the help of the enormous amount of unlabeled target data, which is one of the essential differences between FSL and SSDA.
% \htcomment{
% It is not totally clear you need to get into the FSL literature survey. It opens a can of questions on why you are not taking more sophisticated FSL techniques, but only used ProtoNet. So we can consider avoiding this discussion.
% }

\section{Proposed Framework}
\label{sec:proposed}

Next, we propose a novel SSDA framework, \textit{Source Label Adaptation}. An overview of our proposed framework is shown in Figure~\ref{fig:archi}. In Section~\ref{subsec:danll}, we connect the (SS)DA problem to NLL and point out that a classic NLL method~\cite{reed2014} cannot be directly applied to solve SSDA. In Section~\ref{subsec:ppc}, we review a classic few-shot learning algorithm, \textit{Prototypical Network} \cite{protonet2017} and propose \textit{Protonet with Pseudo Centers} to better estimate the prototypes. In Section~\ref{subsec:lcda}, we summarize our framework and describe the implementation details.
% \htcomment{
% Next, we propose a novel SSDA framework, \textit{Source Label Adaptation}. 
% }
% \htcomment{
% we connect the (SS)DA problem to NLL and point out that a classic NLL method~\cite{} cannot be directly applied to solve SSDA.
% }
% \htcomment{
% Why not present a motivating example trained from source data + ideal target label? I think you can open by this before going into noisy label learning. That also helps you define $g^*$
% }

\subsection{Domain Adaptation as Noisy Label Learning}
\label{subsec:danll}
In Domain Adaptation, we seek an ideal model $g^*$ that can minimize unlabeled target risk.  Ideally, the most suitable label for a source instance $\xx_i^s$ in the target space should be $g^*(\xx_i^s)$. That is, the ideal source loss $\mL_s^*$ is:
\begin{equation}
    \mL_s^*(g | S) = \frac{1}{|S|} \sum_{i=1}^{|S|}H(g(\xx_i^s), g^*(\xx_i^s)),
\end{equation}
where $H$ measures the cross entropy between two distributions.

Combining with the labeled target loss $\mL_\ell$, we refer to the model trained by $\mL_s^*$ and $\mL_\ell$ as \textit{ideally-adapted S+T}. The results of the ideally-adapted S+T reveal the full potential to adapt source labels. As shown in Table~\ref{tab:ideal}, there is a significant difference in performance between a standard S+T and an ideally-adapted S+T, demonstrating that performance can be dramatically affected by only modifying the source labels.
% It hints us that for a source instance $\xx_i^s$ with label $\yy_i^s$, the clean label that best matches the target space should be $g^*(\xx_i^s)$. 

\begin{table}
\centering
\setlength\tabcolsep{7.5 pt}
\scalebox{0.9}{
\begin{tabular}{lcccc}
\toprule
& \multicolumn{2}{c}{A $\to$ C} & \multicolumn{2}{c}{P $\to$ C}  \\
Method           & \small1-shot        & \small3-shot        & \small1-shot        & \small3-shot        \\ \midrule
 S+T  &52.9   & 58.1& 48.8 & 55.5\\
    ideally-adapted S+T &82.9&87.4  &81.6 &86.0\\
\bottomrule

\end{tabular}
}
    \caption{Accuracy (\%) of S+T and ideally-adapted S+T on the 3-shot OfficeHome dataset with ResNet34.
    % In the ideal case, where we have access to the ideal target model, the performance can be dramatically influenced by only modifying the source labels to match the target view. 
    }
    \label{tab:ideal}
\end{table}

In practice, however, we can only approximate the ideal model. To address the issue, we take the original source labels as a noisy version of the ideal labels and approach DA as a NLL problem. We first apply a simple method proposed by \cite{reed2014} to help correct the source labels, which we refer to it as \textit{label correction with self-prediction} \cite{wang2022proselflc}. Specifically, for each source instance $\xx_i^s$, we construct the modified source label $\hyy_i^s$ by combining the original label $\yy_i^s$ and the prediction from the current model $g$ with a ratio $\alpha$.

\begin{equation}
    \hyy_i^s = (1 - \alpha) \cdot \yy_i^s + \alpha \cdot g(\xx_i^s)
    \label{eq:modified_source_label}
\end{equation}

Then, the modified source loss $\hat{\mL}_s$ is:

\begin{equation}
    \hat{\mL}_s(g | S) = \frac{1}{|S|} \sum_{i=1}^{|S|} H(g(\xx_i^s), \hyy_i^s)
\end{equation}
% \htcomment{YOu did not have $\tilde{L}$ nor $L$ before this equation, so no one knows what you mean by "instead"}

% The ideal label correction model should be able to minimize unlabeled target risk in order to perfectly fit the unlabeled data. Practically, we choose a sufficiently good model that can provide insight from target data.

% In this framework, the label correction model in \cite{reed2014} is the model itself, we refer it as label correction with self-prediction:
% \begin{equation}
%     \tyy_i^s = (1 - \alpha) \cdot \yy_i^s + \alpha \cdot g(\xx_i^s)
% \end{equation}

However, in DA, such a method might not be helpful since the model usually overfits the source data, which makes $g(\xx_i^s) \approx \yy_i^s$. That is, the modified source label $\hyy_i^s$ can be almost the same as the original source label $\yy_i^s$ according to Eq.~\ref{eq:modified_source_label}.
% \begin{equation}
%     \begin{aligned}
%         \hyy_i^s &= (1 - \alpha) \cdot \yy_i^s + \alpha \cdot g(\xx_i^s) \\
%         &\approx  (1 - \alpha) \cdot \yy_i^s + \alpha \cdot \yy_i^s
%         = \yy_i^s
%     \end{aligned}
% \end{equation}

\begin{figure}
  \centering
  \includegraphics[width=0.91\linewidth]{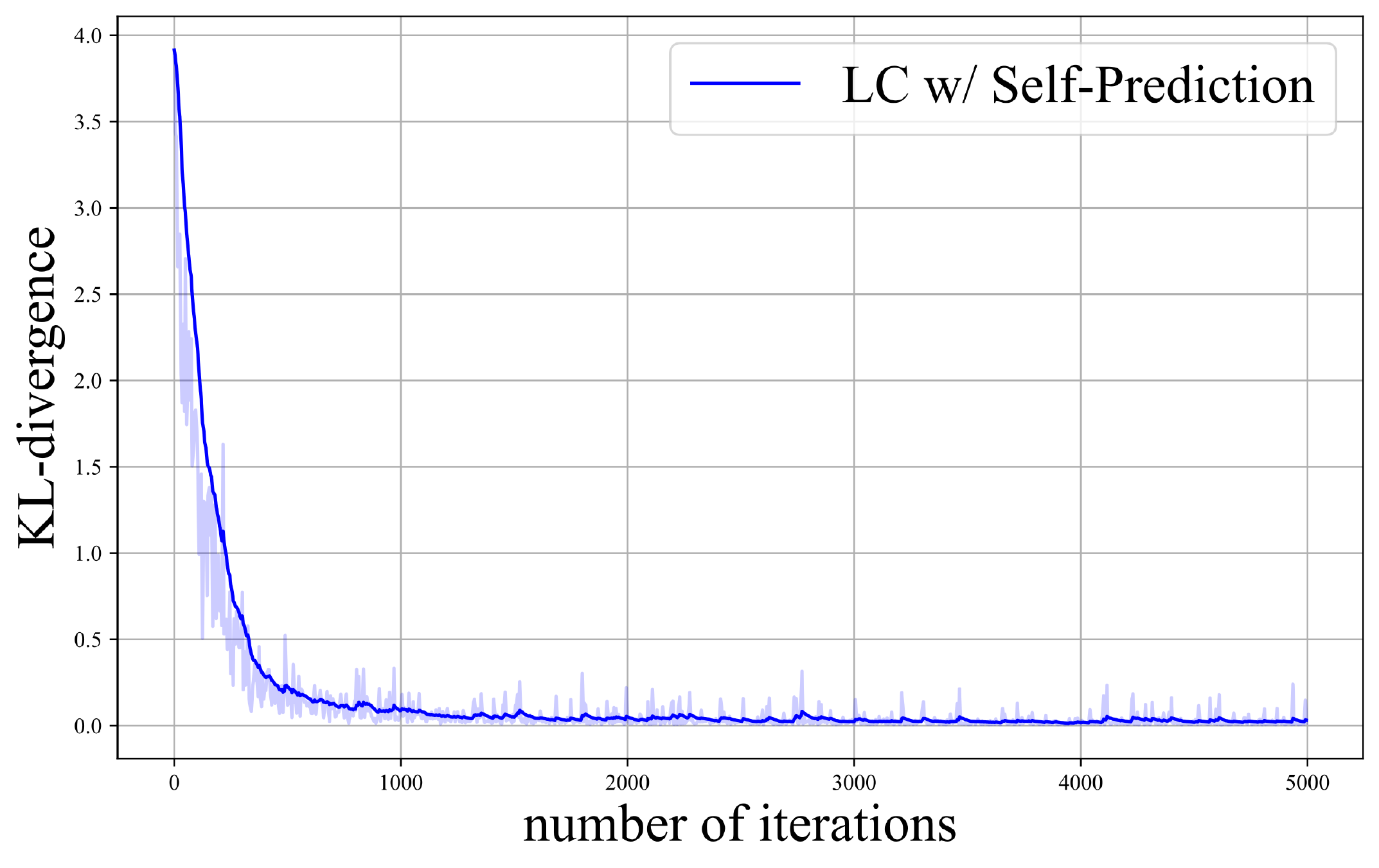}
   \caption{Average KL divergence from $\yy^s$ to $g(\xx^s)$ at each iteration (3-shot \textit{Office-Home} A $\to$ C with ResNet34, smoothing by EMA with a ratio 0.8). }
   
   \label{fig:kl-loss}
\end{figure}

Figure~\ref{fig:kl-loss} shows that when doing \textit{label correction with self-prediction}, the KL divergence from $\yy^s$ to $g(\xx^s)$ could be close to $0$ after 2000 iterations, indicating that the self-prediction is almost the same as the original label. In this case, doing the correction is nearly equivalent to not doing so.

To benefit from the modified labels, we need to eliminate supervision from source data. As an ideal clean label is the output from an ideal model $g^*$, we should instead find a \textit{label adaptation model} $g_c$ that can approximate the ideal model and adapt the source labels to the view of target data. We define an adapted labels $\tyy_i^s$ as a convex combination between the original labels $\yy_i^s$ and the output from $g_c$, which is the same as \cite{reed2014}.

\begin{equation}
    \tyy_i^s = (1 - \alpha) \cdot \yy_i^s + \alpha \cdot g_c(\xx_i^s)
\end{equation}

\subsection{Protonet with Pseudo Centers}
\label{subsec:ppc}
In the semi-supervised setting, we can access a few target labels. Nonetheless, learning from a limited number of target labels might suffer from a severe overfitting issue. Thus, we learn a prototypical network (protonet) \cite{protonet2017} to overcome the few-shot problem.

Given a dataset $\{\xx_i, y_i\}_{i=1}^N$ and a feature extractor $f$; let $N_k$ denote the number of data labeled with $k$. The prototype of class $k$ is defined as the center of features with the same class:

\begin{equation}
    \cc_k = \frac{1}{N_k}\sum_{i=1}^{N} \mathbbm{1}\{y_i = k\} \cdot f(\xx_i).
    \label{eq:center}
\end{equation}

Let $\CC_f = \{\cc_1, \ldots, \cc_K\}$ collects all centers with extractor $f$. We define $P_{\CC_f}: X \mapsto Y$ as a protonet with centers $\CC_f$:

\begin{equation}
    P_{\CC_f}(\xx_i)_k = \frac{\exp(-d(f(\xx_i),\cc_k) \cdot T)}{\sum_{j=1}^K\exp(-d(f(\xx_i), \cc_j) \cdot T)}
    \label{eq:protonet}
\end{equation}

Here $d: F \times F \mapsto [0, \infty)$ is a distance measure over feature space $F$, usually measuring Euclidean distance. $T$ is a hyper-parameter that controls the smoothness of output. As $T \to 0$, the output of a protonet would be close to a uniform distribution.

Since we have access to the labeled target dataset $L$, by Eq.~\ref{eq:center} and Eq.~\ref{eq:protonet}, we can derive labeled target centers $\CC_f^\ell$, and construct a protonet with labeled target centers $P_{\CC_f^\ell}$.

\cite{protonet2017} demonstrated that when $d$ measures Euclidean distance, a protonet is equivalent to a linear classifier with particular parameterization over $F$. Thus, we can take the protonet as a label adaptation model over a particular feature space. The protonet with labeled target centers is purely built from the viewpoint of target data, which should reduce our concerns about the issue mentioned in Section~\ref{subsec:danll}.

However, for a protonet, the ideal centers $\CC_f^*$ should be derived through the unlabeled target dataset $\{(\xx_i^u, y_i^u)\}_{i=1}^{|U|}$. Since we have only a few target labels per class, the labeled target centers $\CC_f^\ell$ are far away from the ideal centers $\CC_f^*$. To better estimate the ideal centers, we propose to find the pseudo centers for unlabeled target data.

With the current model $g$, the pseudo label $\ty_i^u$ for an unlabeled target instance $\xx_i^u$ is:
\begin{equation}
    \ty_i^u = \arg \max_k g(\xx_i^u)_k
    \label{eq:pseudo}
\end{equation}

After deriving unlabeled target data with pseudo labels $\{(\xx_i^u, \ty_i^u)\}_{i=1}^{|U|}$, we can get pseudo centers $\tCC_f$ by Eq.~\ref{eq:center}, and further define a Protonet with Pseudo Centers (PPC) $P_{\tCC_f}$ by Eq.~\ref{eq:protonet}.

Table~\ref{tab:l2-dis} compares the average L2 distance from ideal centers $\CC_f^*$ to labeled target centers $\CC_f^\ell$ and pseudo centers $\tCC_f$ over the feature space trained by S+T. The distance between $\tCC_f$ and $\CC_f^*$ is significantly shorter than the distance between $\CC_f^\ell$ and $\CC_f^*$, which means the pseudo centers are indeed much closer to the ideal centers. 

Taking PPC as the label adaptation model, the modified label $\tyy_i^s$ turns out to be:

\begin{equation}
    \tyy_i^s = (1 - \alpha) \cdot \yy_i^s + \alpha \cdot P_{\tCC_f}(\xx_i^s)
    \label{eq:lc}
\end{equation}

\begin{table}
\centering
\scalebox{0.9}{
    \begin{tabular}{lcc}
    \toprule
    From / To  & labeled target centers & pseudo centers \\ \midrule
    ideal centers & 10.02                  & 4.06  \\
    
    \bottomrule
    \end{tabular}
    }
    \caption{Average L2 Distance from ideal centers to labeled target centers / pseudo centers over the feature space trained by S+T (3-shot \textit{Office-Home} A $\to$ C with ResNet34).}
    \label{tab:l2-dis}
\end{table}

\subsection{Source Label Adaptation for SSDA}
\label{subsec:lcda}
We propose a label adaptation loss for source data to replace the typical source loss with a standard cross entropy loss. For each source instance $\xx_i^s$ with label $\yy_i^s$, we first compute the modified source label $\tyy_i^s$ by Eq.~\ref{eq:lc}. Then, the label adaptation loss $\tilde{\mL}_s$ is:
\begin{equation}
    \tilde{\mL}_s(g|S) = \frac{1}{|S|}\sum_{i=1}^{|S|}H(g(\xx_i^s), \tyy_i^s)
\end{equation}

Our framework, Source Label Adaptation (SLA) for SSDA, can be trained by the following loss function. 
\begin{equation}
    \mL_{\text{SSDA w/ SLA}} = \tilde{\mL}_s + \mL_\ell + \mL_u
\end{equation}
$\mL_\ell$ is the loss function for labeled target data $L$, which can still be a standard cross entropy loss. In contrast to other widely used methods, we primarily concentrate on improving the usage of source data. Therefore, the loss function for unlabeled target data $\mL_u$ can be derived through any state-of-the-art algorithm, and our framework can be easily coupled with other methods without contradiction.

\subsubsection{Implementation Details}

\issue{Warmup Stage.} Our label adaptation framework relies on the quality of the predicted pseudo labels. However, the prediction from the initial model can be noisy. Thus, we introduce a hyperparameter $W$ for warmup to get more stable pseudo labels. During the warmup stage, we train our model normally with original source labels. Specifically, at the $e$-th iteration, we compute the modified source label $\tyy_i^s$ as follows:

\begin{equation}
    \tyy_i^s = \begin{cases}
        \hfil \yy_i^s &\text{if } e \le W\\
        (1 - \alpha) \cdot \yy_i^s + \alpha \cdot P_{\tCC_f}(\xx_i^s) &\text{otherwise}
    \end{cases}
    \label{eq:overall}
\end{equation}
% \htcomment{
% Use $\le$ instead of $<=$. Always try to make your equations look professional.
% }
\issue{Dynamic Update.} The feature space and the predicted pseudo labels constantly evolve during the training phase. By updating the pseudo labels and centers, we can remain the quality of the projected pseudo centers the same. It would be ideal for updating the centers at each iteration. In practice, we update the pseudo labels through Eq.~\ref{eq:pseudo} and update centers with the current feature extractor $f$ through Eq.~\ref{eq:center} for every specific interval $I$. 
Prior works \cite{liang2021domain} have addressed a similar issue and proposed to maintain a memory bank for dynamic updates of the estimated centers. However, in our framework, we need to update both the estimated centers and pseudo labels simultaneously. Therefore, we decided to adopt a more straightforward solution to mitigate the demands on time and complexity.

\section{Experiments}
\label{sec:exp}

\begin{table*}[ht]
\centering
\scalebox{0.76}{
\begin{tabular}{lcccccccccccccccc}
        \toprule
        & \multicolumn{2}{c}{R $\rightarrow$ C} 
        & \multicolumn{2}{c}{R $\rightarrow$ P} 
        & \multicolumn{2}{c}{P $\rightarrow$ C}
        & \multicolumn{2}{c}{C $\rightarrow$ S} 
        & \multicolumn{2}{c}{S $\rightarrow$ P}
        & \multicolumn{2}{c}{R $\rightarrow$ S} 
        & \multicolumn{2}{c}{P $\rightarrow$ R}
        & \multicolumn{2}{c}{\textbf{Mean}} \\  
        \textbf{Method}
        & \small 1-shot & \small 3-shot
        & \small 1-shot & \small 3-shot
        & \small 1-shot & \small 3-shot
        & \small 1-shot & \small 3-shot
        & \small 1-shot & \small 3-shot
        & \small 1-shot & \small 3-shot
        & \small 1-shot & \small 3-shot
        & \small \textbf{1-shot} & \small \textbf{3-shot}\\
        \midrule
S+T &55.6 &60.0 &60.6 &62.2 &56.8 &59.4 &50.8 &55.0 &56.0 &59.5 &46.3 &50.1 &71.8 &73.9 &56.9 &60.0\\
DANN \cite{dann2015} &58.2 &59.8 &61.4 &62.8 &56.3 &59.6 &52.8 &55.4 &57.4 &59.9 &52.2 &54.9 &70.3 &72.2 &58.4 &60.7\\
ENT \cite{ent2004} &65.2 &71.0 &65.9 &69.2 &65.4 &71.1 &54.6 &60.0 &59.7 &62.1 &52.1 &61.1 &75.0 &78.6 &62.6 &67.6\\
APE \cite{ape2020} &70.4 &76.6 &70.8 &72.1 &72.9 &76.7 &56.7 &63.1 &64.5 &66.1 &63.0 &67.8 &76.6 &79.4 &67.6 &71.7 \\
DECOTA \cite{decota2020} &79.1 &80.4 &74.9 &75.2 &76.9 &78.7 &65.1 &68.6 &{72.0} &72.7 &69.7 &71.9 &79.6 &81.5 &73.9 &75.6 \\
MCL \cite{mcl2022} &77.4 &79.4 &74.6 &\textbf{76.3} &75.5& 78.8 &66.4 &70.9 &\textbf{74.0}& \textbf{74.7}& 70.7 &72.3 &\textbf{82.0} &\textbf{83.3}& 74.4 &76.5\\
\hline
MME \cite{mme2019}              & 70.0          & 72.2          & 67.7          & 69.7          & 69.0          & 71.7          & 56.3          & 61.8          & 64.8          & 66.8          & 61.0          & 61.9          & 76.1           & 78.5          & 66.4          & 68.9          \\
MME + SLA (ours)  & 71.8          & 73.3          & 68.2          & 70.1          & 70.4          & 72.7          & 59.3          & 63.4          & 64.9          & 67.3          & 61.8          & 63.9          & 77.2           & 79.6          & 68.8          & 70.0          \\\hline
CDAC \cite{cdac2021}             & 77.4          & 79.6          & 74.2          & 75.1          & 75.5          & 79.3          & 67.6          & 69.9          & 71.0          & 73.4 & 69.2          & 72.5          & {80.4}  & 81.9          & 73.6          & 76.0          \\ 

CDAC + SLA (ours) & \textbf{79.8} & \textbf{81.6} & \textbf{75.6} & {76.0} & \textbf{77.4} & \textbf{80.3} & \textbf{68.1} & \textbf{71.3} & {71.7} & {73.5}          & \textbf{71.7} & \textbf{73.5} & {80.4}  & {82.5} & \textbf{75.0} & \textbf{76.9} \\ \bottomrule
% \end{tabularx}
\end{tabular}
}
\caption{Accuracy (\%) on \textit{DomainNet} for 1-shot and 3-shot Semi-Supervised Domain Adaptation (ResNet34).}
\label{tab:domainnet}
\end{table*}

\begin{table*}[ht]
\centering
\setlength\tabcolsep{9.5 pt}
\scalebox{0.76}{
\begin{tabular}{lccccccccccccc}
\toprule
\textbf{Method} & A$\to$C & A$\to$P & A$\to$R & C$\to$A & C$\to$P & C$\to$R & P$\to$A & P$\to$C & P$\to$R & R$\to$A & R$\to$C & R$\to$P   & \textbf{Mean} \\ \midrule
\multicolumn{14}{c}{\textbf{One-shot}}\\
\midrule
S+T & 50.9 & 69.8 & 73.8 & 56.3 & 68.1 & 70.0 & 57.2 & 48.3 & 74.4 & 66.2 & 52.1 & 78.6 & 63.8 \\ 
DANN \cite{dann2015} & 52.3 & 67.9 & 73.9 & 54.1 & 66.8 & 69.2 & 55.7 & 51.9 & 68.4 & 64.5 & 53.1 & 74.8 & 62.7 \\ 
ENT \cite{ent2004} & 52.9 & 75.0 & 76.7 & 63.2 & 73.6 & 73.2 & 63.0 & 51.9 & 79.9 & 70.4 & 53.6 & 81.9 & 67.9 \\
APE \cite{ape2020}& 53.9&76.1&75.2&63.6&69.8&72.3&63.6&58.3&78.6&72.5&60.7&81.6&68.9\\
DECOTA\cite{decota2020} &42.1 &68.5&72.6&60.3&70.4&70.7&60.0&48.8&76.9&71.3&56.0&79.4&64.8\\\midrule
MME \cite{mme2019}& 59.6 & 75.5 & 77.8 & 65.7 & 74.5 & 74.8 & 64.7 & 57.4 & 79.2 & 71.2 & 61.9 & 82.8 & 70.4 \\ 
MME + SLA (ours) & 62.1 & 76.3 & 78.6 & \textbf{67.5} & 77.1 & 75.1 & 66.7 & 59.9 & 80.0 & \textbf{72.9} & 64.1 & 83.8 & 72.0 \\  \hline
CDAC \cite{cdac2021} & 61.2 & 75.9 & 78.5 & 64.5 & 75.1 & 75.3 & 64.6 & 59.3 & 80.0 & 72.7 & 61.9 & 83.1 & 71.0 \\
CDAC + SLA (ours) & \textbf{63.0} & \textbf{78.0} & \textbf{79.2} & 66.9 & \textbf{77.6} & \textbf{77.0} & \textbf{67.3} & \textbf{61.8} & \textbf{80.5} & 72.7 & \textbf{66.1} & \textbf{84.6} & \textbf{72.9} \\ 

\midrule
\multicolumn{14}{c}{\textbf{Three-shot}}\\
\midrule
S+T & 54.0 & 73.1 & 74.2 & 57.6 & 72.3 & 68.3 & 63.5 & 53.8 & 73.1 & 67.8 & 55.7 & 80.8 & 66.2 \\ 
DANN \cite{dann2015} & 54.7 & 68.3 & 73.8 & 55.1 & 67.5 & 67.1 & 56.6 & 51.8 & 69.2 & 65.2 & 57.3 & 75.5 & 63.5 \\ 
ENT \cite{ent2004}& 61.3 & 79.5 & 79.1 & 64.7 & 79.1 & 76.4 & 63.9 & 60.5 & 79.9 & 70.2 & 62.6 & 85.7 & 71.9\\
APE \cite{ape2020} &63.9&81.1&80.2&66.6&79.9&76.8&66.1&65.2&82.0&73.4&66.4&86.2&74.0\\
DECOTA \cite{decota2020} &64.0&81.8&80.5&68.0&\textbf{83.2}&79.0&69.9&68.0&82.1&74.0&\textbf{70.4}&\textbf{87.7}&75.7\\
\hline
MME \cite{mme2019} & 63.6 & 79.0 & 79.7 & 67.2 & 79.3 & 76.6 & 65.5 & 64.6 & 80.1 & 71.3 & 64.6 & 85.5 & 73.1 \\ 
MME + SLA (ours) & 65.9 & 81.1 & 80.5 & \textbf{69.2} & 81.9 & 79.4 & 69.7 & 67.4 & 81.9 & \textbf{74.7} & 68.4 & {87.4} & 75.6 \\ \hline
CDAC \cite{cdac2021} & 65.9 & 80.3 & 80.6 & 67.4 & 81.4 & \textbf{80.2} & 67.5 & 67.0 & 81.9 & 72.2 & 67.8 & 85.6 & 74.8 \\ 

CDAC + SLA (ours) & \textbf{67.3} & \textbf{82.6} & \textbf{81.4} & \textbf{69.2} & {82.1} & 80.1 & \textbf{70.1} & \textbf{69.3} & \textbf{82.5} & 73.9 & {70.1} & 87.1 & \textbf{76.3} \\
\bottomrule
\end{tabular}
}
\caption{Accuracy (\%) on \textit{Office-Home} for 1-shot and 3-shot Semi-Supervised Domain Adaptation (ResNet34).}
\label{tab:officehome}
\end{table*}

\begin{table*}
\setlength\tabcolsep{9.5 pt}
\scalebox{0.76}{
\begin{tabular}{lccccccccccccc}
\toprule
Method & A$\to$C & A$\to$P & A$\to$R & C$\to$A & C$\to$P & C$\to$R & P$\to$A & P$\to$C & P$\to$R & R$\to$A & R$\to$C & R$\to$P   & \textbf{Mean} \\ \midrule
MCL \cite{mcl2022} & \textbf{67.5} & \textbf{83.9}&\textbf{82.4}&\textbf{71.4}&\textbf{84.3}&\textbf{81.6}&69.9&\textbf{68.0}&\textbf{83.0}&75.3&\textbf{70.1}&\textbf{88.1} &\textbf{77.1}\\\midrule
MCL* & 64.1 & {81.6} & 80.6 & {70.3} & 82.2 & 79.2 & 70.6 & 64.0 & 81.8 & 75.3 & 67.8 & 86.6 & 75.3 \\ 
MCL + SLA (ours) & 64.3 & 81.6 & 80.8 & 70.2 & {82.6} & {79.4} & \textbf{70.9} & {64.2} & {82.2} & \textbf{75.5} & {68.0} & {86.8} & {75.6} \\ 

\bottomrule

\multicolumn{14}{l}{*: Reproduced by ourselves}
\end{tabular}
}
\caption{Accuracy (\%) of MCL and MCL + SLA on \textit{Office-Home} for 3-shot Semi Supervised Domain Adaptation (ResNet34).}
\label{tab:mcl}
\end{table*}

We first sketch our experiment setup, including data sets, competing methods, and parameter settings in Section~\ref{subsec:expset}.
% \htcomment{
% We first sketch our experiment setup, including data sets, competing methods, and parameter settings in Subsection~\ref{subsec:expset}.
% }
We then present experimental results to validate the superiority of the proposed SLA framework in Section~\ref{subsec:sota}.
% \htcomment{
% We then present experimental results to validate the superiority of the proposed SLA framework. 
% }
We further analyze our proposed framework and highlight the limitation in Section~\ref{subsec:pl}.

\subsection{Experiment Setup}
\label{subsec:expset}
\issue{Datasets.} We evaluate our proposed SLA framework on two sets of SSDA benchmarks, including \textit{Office-Home} \cite{venkateswara2017deep} and \textit{DomainNet} \cite{peng2019moment}. 
% \htcomment{
% If you have started using acronyms, it is generally best to stick to SLA instead of changing back and forth.
% }
% \htcomment{
% "several" usually means 3-5. So if you only have two (sets), say two sets of benchmarks.
% }
\textit{Office-Home} is a mainstream benchmark for both UDA and SSDA. It contains four domains: Art (A), Clipart (C), Product (P), and Real (R), with 65 categories. \textit{DomainNet} is initially designed for benchmarking Multi-Source Domain Adaptation approaches. \cite{mme2019} pickup four domains: Real (R), Clipart (C), Painting (P), and Sketch (S) with 126 classes to build a cleaner dataset for SSDA. Besides, they focus on seven scenarios instead of combining all pairs. Our experiments follow the settings in recent works \cite{mme2019, cdac2021, mcl2022}, with the same sampling strategy for both the training set and validation set, and we conduct both 1-shot and 3-shot settings on all datasets.

% \begin{figure*}[ht]
%     \centering
%     \includegraphics[width=\linewidth]{Imgs/exp_prob.eps}
%     \caption{Average adapted source labels from the PPC and ideal S+T for a certain class (3-shot Office-Home A $\to$ C with ResNet34). x-axis: the classes, y-axis: the probability of the average adapted labels. We illustrate the average adapted source labels in S+T + SLA on six representative classes. Note that the original source labels should be one-hot encoded. The results show that the adapted labels could be much closer to the ideal labels.}
%     \label{fig:prob}
% \end{figure*}

\issue{Implementation Details.} Our framework can be applied with many state-of-the-art methods, we choose MME \cite{mme2019}, and CDAC \cite{cdac2021} as our cooperators to validate the efficacy of our method, named MME + SLA and CDAC + SLA, respectively. For a fair comparison, we choose ResNet34 \cite{resnet2015} as our backbone. The backbone is pre-trained on ImageNet-1K dataset \cite{imagenet}, and the model architecture, batch size, learning rate scheduler, optimizer, weight-decay, and initialization strategy are all followed as previous works \cite{mme2019, cdac2021, mcl2022}. We follow the same hyper-parameters for MME and CDAC as their suggestions. We set the mix ratio $\alpha$ in Eq.~\ref{eq:overall} to $0.3$ and the temperature parameter $T$ in Eq.~\ref{eq:protonet} to $0.6$. The update interval $I$ mentioned in Section~\ref{subsec:lcda} is $500$. The warmup parameter $W$ in Eq.~\ref{eq:overall} is $500$ for MME on \textit{Office-Home}; $2000$ for CDAC on \textit{Office-Home}; $3000$ for MME on \textit{DomainNet}; $50000$ for CDAC on \textit{DomainNet}. After the warmup stage, we refresh the learning rate scheduler so that the label adaptation loss can be updated with a higher learning rate. All hyper-parameters can be properly tuned via the validation process. For each subtask, we conducted the experiments three times. The detailed statistics of our results can be found in our supplementary materials.

\subsection{Comparison with State-of-the-Arts}
\label{subsec:sota}
We compare our results with several baselines, including \textbf{S+T}, \textbf{DANN} \cite{dann2015}, \textbf{ENT} \cite{ent2004}, \textbf{MME} \cite{mme2019} \textbf{APE} \cite{ape2020}, \textbf{CDAC} \cite{cdac2021}, \textbf{DECOTA} \cite{decota2020}, \textbf{MCL} \cite{mcl2022}. \textbf{S+T} is a baseline method for SSDA, with only source data and labeled target data involved in the training process. \textbf{DANN} is a classic unsupervised domain adaptation method, and \cite{mme2019} reproduces it by training with additional labeled target data. \textbf{ENT} is a standard entropy minimization originally designed for Semi-Supervised Learning, and the reproduction was also done by \cite{mme2019}. Note that for MCL, we only compare with their results on \textit{DomainNet}. We leave the detailed analysis for MCL on \textit{Office-Home} in Section~\ref{subsec:pl}.

\issue{\textit{DomainNet}.} We show the results on \textit{DomainNet} dataset with 1-shot and 3-shot settings on Table~\ref{tab:domainnet}. It is worth noting two things. First, for MME and CDAC, almost all sub-tasks get improvement after applying our SLA framework, except for only two cases where CDAC + SLA performs roughly the same as CDAC. Second, the overall performance of CDAC + SLA for 1-shot and 3-shot settings reaches 75.0\% and 76.9\%, respectively; both outperform the previous methods and achieve new state-of-the-art results.

% \htcomment{
% People would ask what the error bar is. Any good answers?
% }
% \htcomment{
% Maybe put MME and MME+SLA as the same group, and CDAC and CDAC+SLA as another group. That highlights the message better.
% }
\issue{\textit{Office-Home}.} We show the results on \textit{Office-Home} dataset with 1-shot and 3-shot settings on Table~\ref{tab:officehome}. Similarly, after applying SLA to MME and CDAC, the performances get much better except for only one case under the 3-shot setting. Overall, our framework improves the original works by at least 1.5\% under all settings.

\subsection{Analysis}
\label{subsec:pl}
\issue{Reproducibility issue for MCL.} MCL \cite{mcl2022} designs consistency regularization for SSDA at three different levels and achieves excellent results. However, our experiments cannot fully reproduce their reported numbers. The reproduced results on 3-shot \textit{Office-Home} dataset are shown in Table~\ref{tab:mcl}. After applying our SLA framework, although we can stably improve our reproduction, we are still unable to compete with their reported values. We put our detailed reproducing results into the supplementary materials, and the implementation is avaiable at \url{https://github.com/chester256/MCL}.
% \textbf{Comparison with pseudo-labeling strategy.} Our method can be viewed as a pseudo-label-oriented method. 

% \begin{table}
% \centering
%     \begin{adjustbox}{max width=\textwidth}
% \begin{tabular}{c|cccc}
% \hline
% Method &A $\to$ P & C $\to$ A & P $\to$ A & R $\to$ C \\\hline
%  S+T  &74.7 &56.3 &58.1 &59.1\\
%  S+T + PPC &77.1 &59.8 &60.9 & 62.1\\
%  S+T + SLA &\textbf{77.7} &\textbf{60.5} & \textbf{61.3} &\textbf{62.5}\\
% \hline

% \end{tabular}
% \end{adjustbox}
%     \caption{Accuracy (\%) of S+T, S+T + PPC, and S+T + SLA on the 3-shot \textit{Office-Home} with ResNet34. Although directly apply PPC to S+T can boost the performance, we show that learning from the labels modified by PPC can perform much better. }
%     \label{tab:ppc}
% \end{table}

% \issue{PPC for inference.} In SLA, we build a PPC to provide the view from the target data. PPC can be viewed as a variant of the pseudo-labeling method proposed in \cite{shot2021}. They apply such a method to boost their final performance in their work. If PPC has performed well, a natural question is: \textit{Is it necessary to first modify the source labels by PPC, then instead learn from the modified labels?} As shown in Table~ \ref{tab:ppc}, S+T + SLA can perform better than directly taking PPC for inference. It also validates that we can do much better by carefully revisiting the usage of source data.

\begin{figure}[t]
    \centering
    \includegraphics[width=0.91\linewidth]{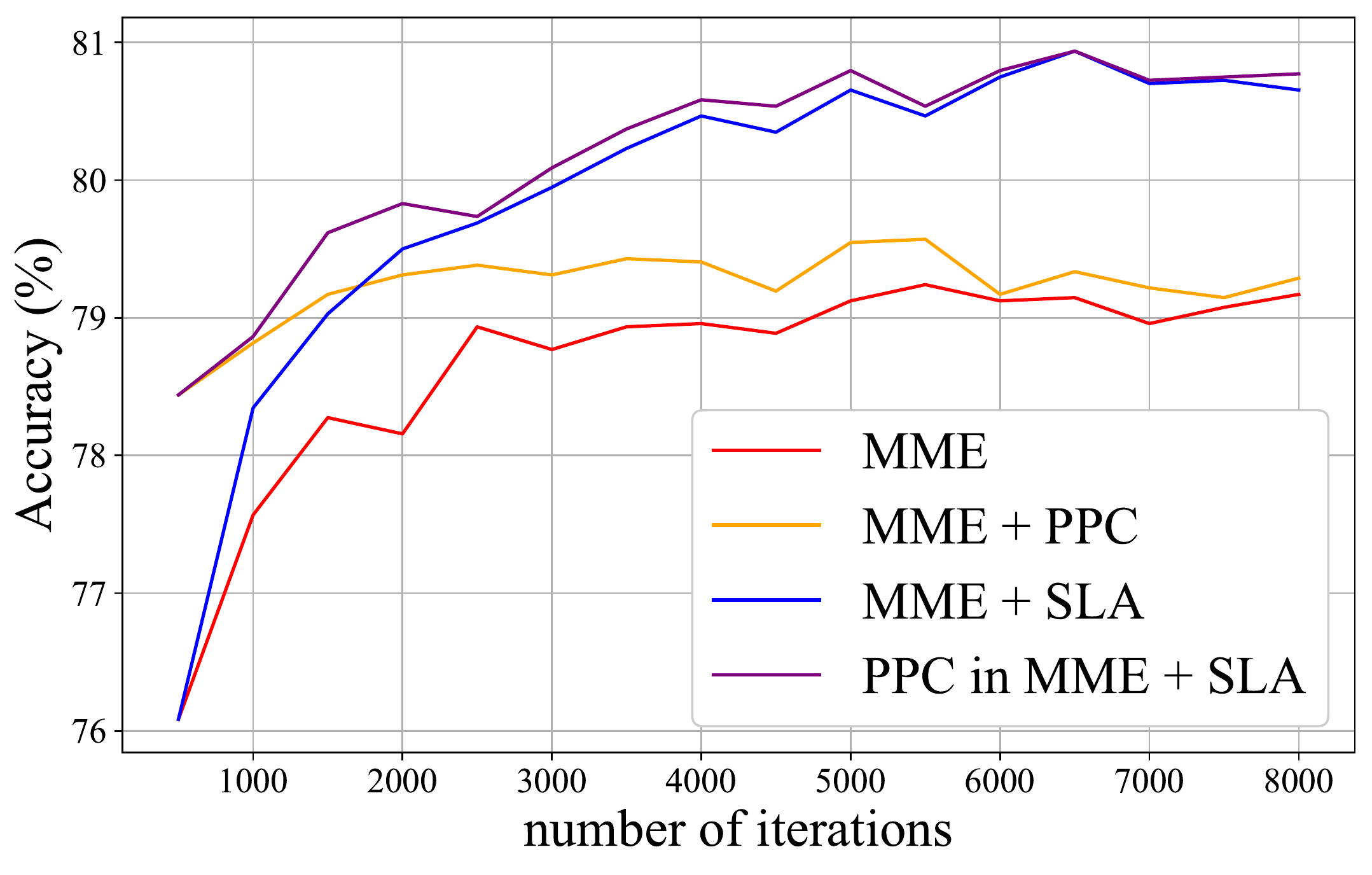}
    \caption{The intermediate results in SLA on 3-shot  \textit{Office-Home} A $\to$ P with ResNet34.}
     \label{fig:intermediate}
\end{figure}

\issue{The intermediate results in SLA.} In SLA, we build a PPC to provide the view from the target data. PPC can be viewed as a variant of the pseudo-labeling method proposed in \cite{shot2021}. They apply such a method to boost their final performance in their work. If PPC has performed well, a natural question is: \textit{Is it necessary to modify source labels by PPC?} 

To reveal the intermediate steps within SLA, we plot the test performance of MME (red), MME + PPC (orange), MME + SLA (blue), and PPC within MME + SLA (purple) during the training phase in Figure~\ref{fig:intermediate}. Initially, PPC (target view) performs at a higher level. However, if the source labels are not adapted, it will end up converge to the same performance as MME. In contrast, within our SLA framework, the model leverages the benefits of PPC, further producing an enhanced version of PPC, resulting in better overall performance compared to the original MME.

% Through the dynamic update described in Section~\ref{subsec:lcda}, our model continuously learns from PPC, and keeps producing an improved PPC, resulting in better overall performance compared to the original MME.

\begin{table}
\centering
\scalebox{0.9}{
% \begin{adjustbox}{width=\linewidth}
\begin{tabular}{lcccccc}
\toprule
$\alpha$   & 0.1   & 0.3 & 0.5   & 0.7   & 0.9   \\ \midrule validation & 74.87 & \textbf{75.94} & 75.76 & 75.38 & 72.39 \\
test       & 74.43 & \textbf{75.57} & 75.38 & 74.79 & 72.76\\

\bottomrule
\end{tabular}
}
% \end{adjustbox}
%\caption{Average validation and test accuracy (\%) on 3-shot \textit{Office-Home} dataset with ResNet34.}
\caption{Average accuracy (\%) of MME + SLA on 3-shot \textit{Office-Home} with ResNet34.}
\label{tab:alpha}
\end{table}

\issue{Sensitivity study of $\alpha$.} Table~\ref{tab:alpha} shows the sensitivity study results for $\alpha$ using MME + SLA on 3-shot \textit{OfficeHome} dataset.
We selected $0.3$ from the best validation performance and kept $0.3$ throughout all experiments for simplicity and resource saving.
% We selected 0.3 throughout all experiments for simplicity and resource saving.
In the mid-range, $\alpha = 0.3$ (favoring source view) or $\alpha = 0.7$ (favoring target view) perform similarly, hinting that it is stable enough for a proper range of choices. It is worth noting that the dynamic adjustment of $\alpha$ is a promising direction. 
We proposed the warmup stage that changes $\alpha$ from $0$ to the desired value after a period of warming up.
More sophisticated scheduling techniques, such as the linear growth approach \cite{kim2021self}, could potentially replace the warmup parameter and lead to further improvements. We leave exploration of these techniques for future work.

\begin{figure}[ht]
  \begin{minipage}{0.09\textwidth}
    \centering
    \includegraphics[width=\textwidth]{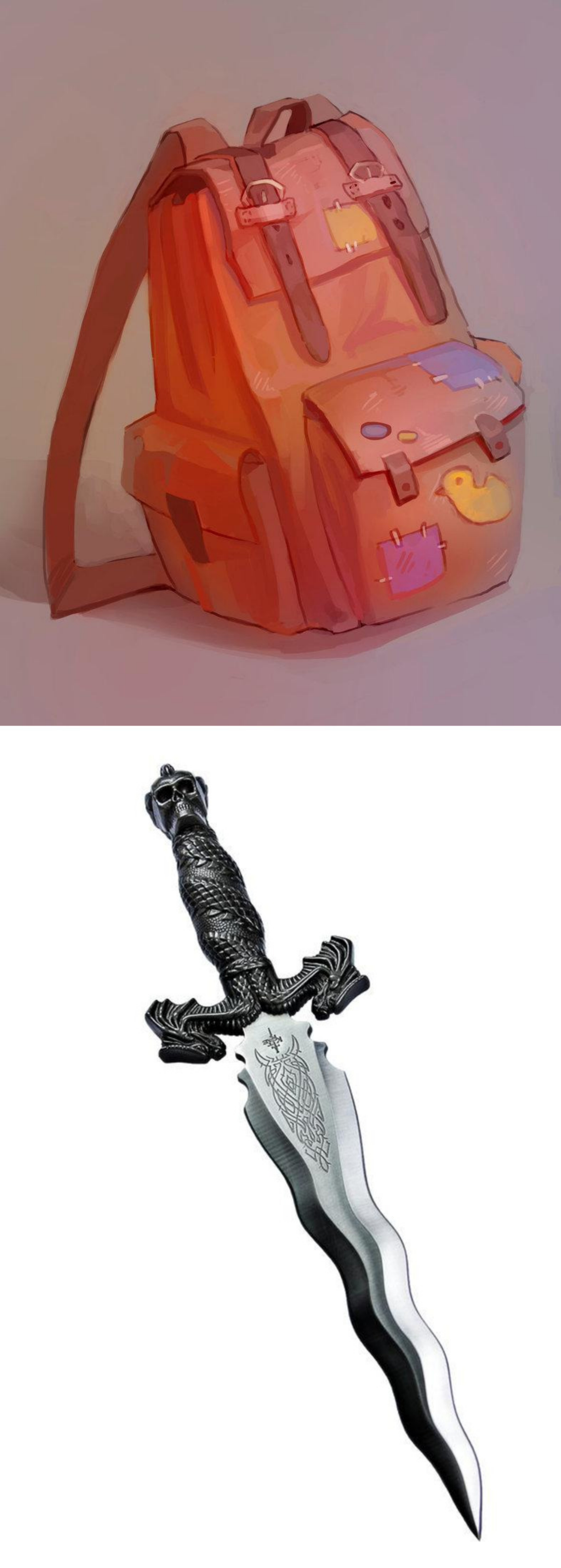}
    \label{fig:subfig1}
  \end{minipage}
  \begin{minipage}{0.35\textwidth}
    \centering
    \includegraphics[width=\textwidth]{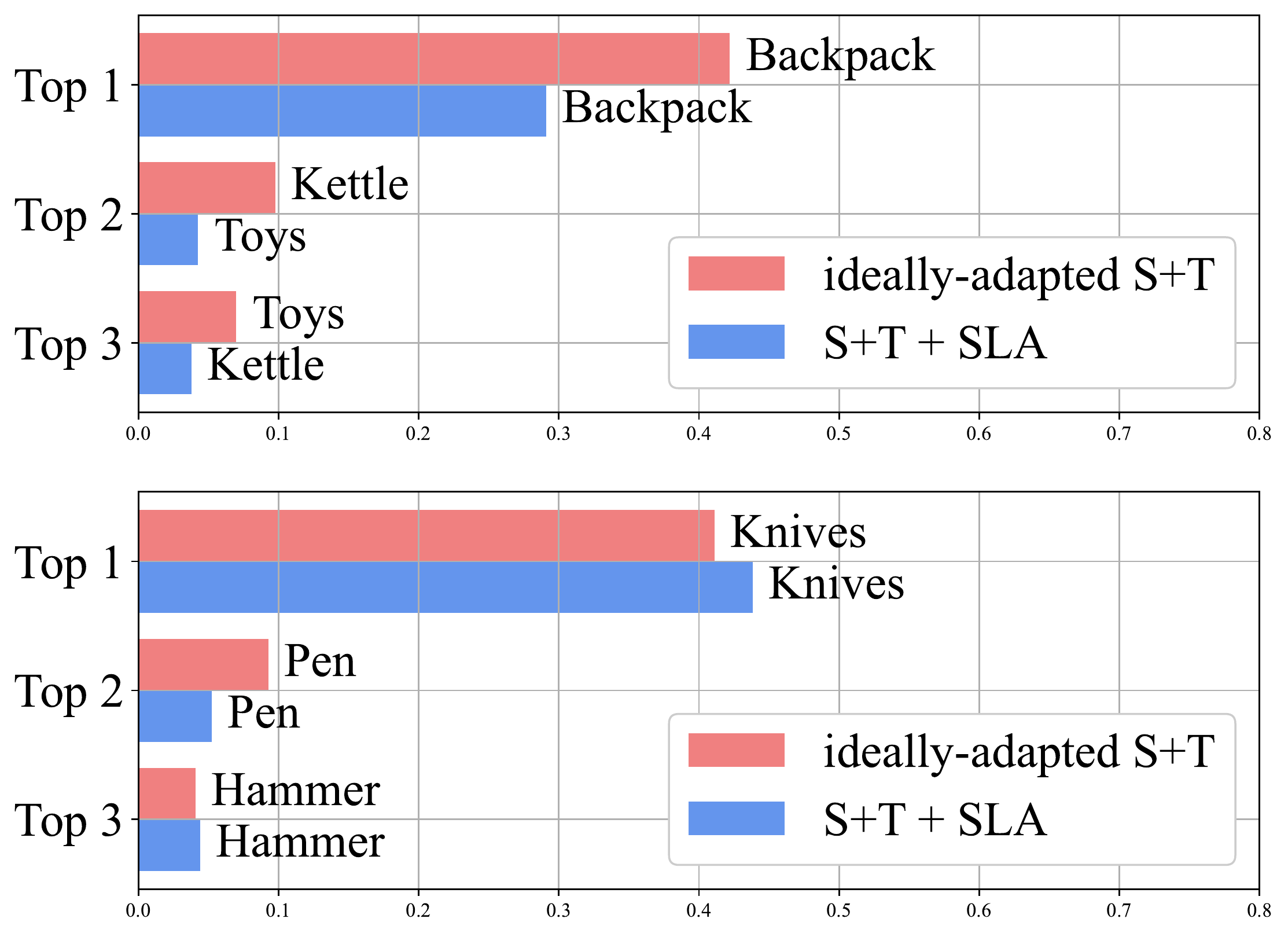}
    \label{fig:subfig2}
  \end{minipage}
  \caption{Top-3 probabilities of the average adapted source labels on 3-shot \textit{Office-Home} A $\to$ C with ResNet34. \textbf{Top.} Class 1 (backpack). \textbf{Bottom.} Class 30 (knives).}
  \label{fig:prob}
\end{figure}

\issue{Illustration of the adapted labels.} We aim to adapt the original source label $\yy_i^s$ to the ideal label $g^*(\xx_i^s)$. To demonstrate the change of labels, we visualize the top 3 probabilities of the average adapted source labels on two classes in Figure~\ref{fig:prob}. For Backpack (class 1), ideally-adapted S+T should change the source label to 40\% Backpack + 10\% Kettle + 8\% Toys; SLA proposes to change to 30\% Backpack + 5\% Toys + 4\% Kettle, which is closer to the ideally-adapted label than the original label of 100\% Backpack. We draw the same conclusion for knives (class 30).
% illustrate the average probability distribution of the adapted labels. The results are shown in Figure~\ref{fig:prob}. Compared with the original source labels, which are one-hot encoded, our adapted labels can be much closer to the ideal labels.

% visualize one example with the top 3 probabilities of the average adapted source label in Fig.~\ref{fig:output}. Ideally-adapted S+T should change the source label to 40\% Backpack + 10\% Kettle + 8\% Toys because the example resembles some kettle/toy in the target domain; SLA proposes to change to 30\% Backpack + 5\% Toys + 4\% Kettle, which is closer to the ideally-adapted label than the original label of 100\% Backpack.

\begin{figure}[t]
    \centering
    \includegraphics[width=0.91\linewidth]{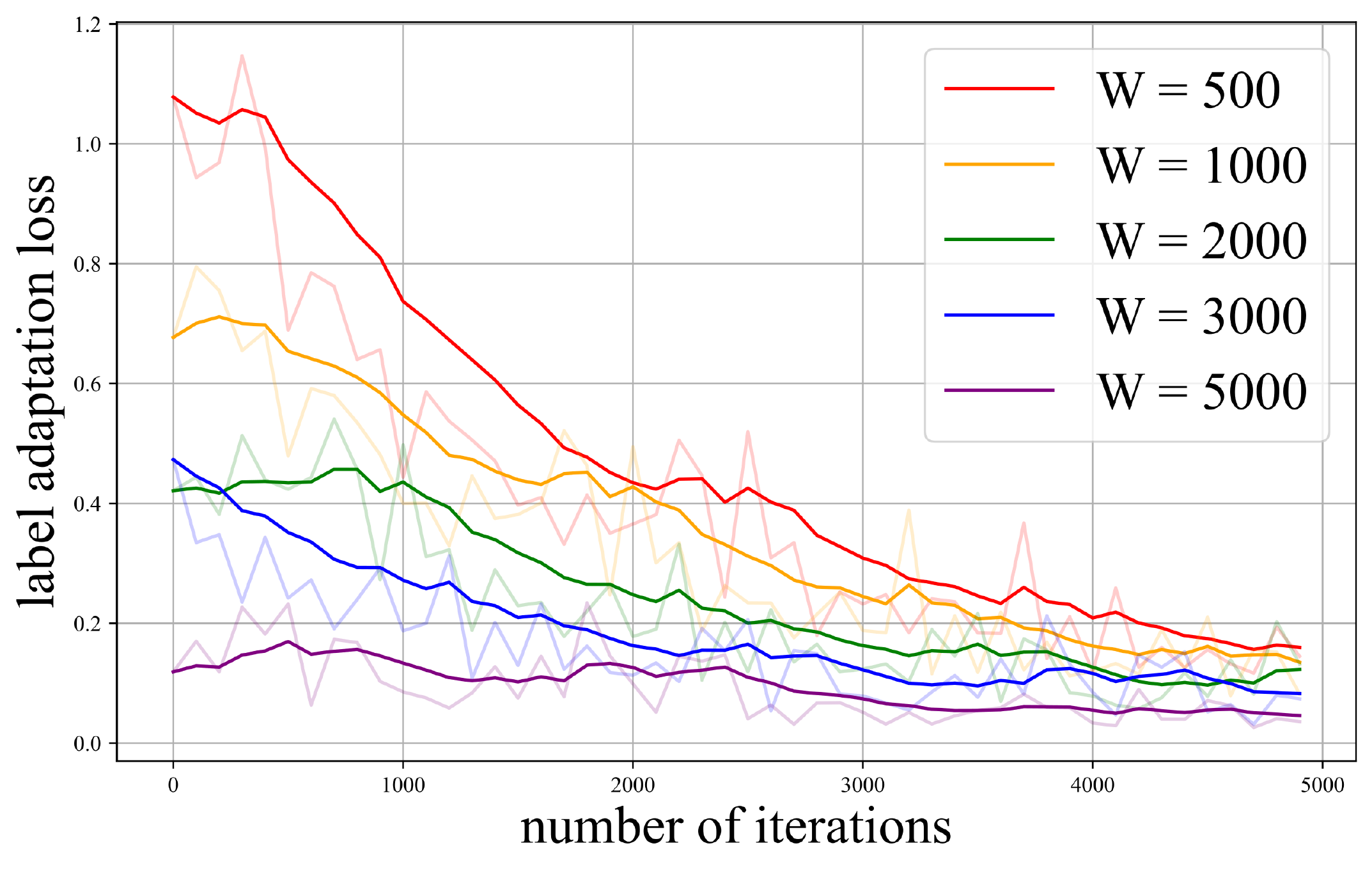}
    \caption{Label Adaptation Loss of MME + SLA by first pre-training MME for $W$ iterations on 3-shot \textit{Office-Home} A $\to$ C with ResNet34. (Smoothing by EMA with a ratio $0.8$.)}
    \label{fig:source_loss}
\end{figure}

\issue{Warmup issue for MME + SLA.} As described in Section~\ref{subsec:lcda}, our framework relies on the quality of the predicted pseudo labels. Thus, we introduce a warmup stage parameter $W$ to derive a robust model. We can treat the warmup strategy as a two-stage algorithm. Take MME as our backbone method; the algorithm works like:
\begin{enumerate}
    \item Train a model with normal MME loss for $W$ iteration.
    \item Take the model above as a pre-trained model and further applying label adaptation loss.
\end{enumerate}
For the first step, intuitively, we should train the model until the loss converges. That is how we select the warmup stage parameter for CDAC + SLA. However, empirically we found that the performance of MME + SLA will degrade if we train an MME model until it converges. Table~\ref{tab:warmup} shows the sensitivity test of $W$ using MME + SLA on \textit{Office-Home} dataset. We can observe that no matter the 1-shot or 3-shot settings, the performance is generally getting worse as the number of warmup stages increases. To analyze the effect, we first pre-train a normal MME for $W$ iterations, then observe the label adaptation loss of MME + SLA. Figure~\ref{fig:source_loss} plots the label adaptation loss of MME + SLA by first pre-training MME for $W$ iterations. We can observe that when $W = 5000$, the initial label adaptation loss has already been close to $0$. Doing label adaptation in the situation is almost equivalent to not doing so, as we mentioned in Section~\ref{subsec:danll}.

\begin{table}[t]
\centering
\setlength\tabcolsep{15 pt}
\scalebox{0.9}{
\begin{tabular}{lcc}
\toprule
                   & \multicolumn{2}{c}{A $\to$ C} \\
Warmup Stage ($W$) & \small1-shot        & \small3-shot        \\ \midrule
500                & \textbf{62.09}         & \textbf{65.90}         \\
1000               & 61.95         & 64.99         \\
2000               & 61.37         & 64.72         \\
3000               & 61.53         & 64.87         \\
5000               & 61.79         & 64.68         \\ \bottomrule
\end{tabular}
}
\caption{Accuracy (\%) for different warmup stage $W$ of MME + SLA on \textit{Office-Home} A $\to$ C with ResNet34.}
\label{tab:warmup}
\end{table}

\issue{Limitation.} Our SLA framework might not be helpful if the label adaptation loss approaches $0$. Although we have applied the Protonet with Pseudo Centers to avoid the issue, the loss will converge to $0$ in MME + SLA. We leave the analysis of the reason for the convergence as a future work. On the other hand, we argue that it is unnecessary to discuss the reason in our proposed scope since we can make a trade-off by carefully tuning the warmup parameter $W$, and the issue turns out to be part of the hyper-parameters selection.

\section{Conclusion}
\label{sec:conclusion}

In this work, we present a general framework, \textit{Source Label Adaptation} for Semi-Supervised Domain Adaptation. Our work highlights that the usage of source data should be revisited carefully. We argue that the original source labels might be noisy from the perspective of target data. We approach Domain Adaptation as a Noisy Label Learning problem and correct source labels with the predictions from Protonet with Pseudo Centers. Our approach mainly addresses an issue that is orthogonal to other existing works, which focus on improving the usage of unlabeled data. The empirical results show that we can apply our framework to several state-of-the-art algorithms for SSDA and further boost their performances.

\issue{Acknowledgement.} We thank the anonymous reviewers for valuable comments. This work is supported by the Ministry of Science and Technology in Taiwan via the grant MOST 111-2628-E-002-018. We thank to National Center for High-performance Computing (NCHC) in Taiwan for providing computational and storage resources.

% \section{Table}
% \input{Section/Table}

% \input{Section/Second}

%%%%%%%%% REFERENCES
{\small
\bibliographystyle{ieee_fullname}
\bibliography{egbib}
}

\end{document}